\documentclass[10pt,letterpaper]{article}

\usepackage{iccv}
\usepackage{times}
\usepackage{epsfig}
\usepackage{graphicx}
\usepackage{amsmath}
\usepackage{amssymb}
\usepackage{makecell}
\usepackage{multirow}
\usepackage{subcaption} 
\usepackage{bbm}
\usepackage{authblk}

\usepackage[pagebackref=true,breaklinks=true,letterpaper=true,colorlinks,bookmarks=false]{hyperref}

\iccvfinalcopy 

\ificcvfinal\fi
\begin{document}

\title{Label Super Resolution with Inter-Instance Loss}
\author[1]{Maozheng Zhao}
\author[1]{Le Hou}
\author[1]{Han Le}
\author[1]{Dimitris Samaras}
\author[2]{Nebojsa Jojic}
\author[1]{Danielle Fassler}
\author[1]{Tahsin Kurc}
\author[1]{Rajarsi Gupta}
\author[2,3]{Kolya Malkin}
\author[1]{Shroyer Kenneth}
\author[1]{Joel Saltz}
\affil[1]{Stony Brook University}
\affil[2]{Microsoft Research}
\affil[3]{Yale University}

\maketitle

\begin{abstract}

    
    For the task of semantic segmentation, high-resolution (pixel-level) ground truth is very expensive to collect, especially for high resolution images such as gigapixel pathology images. On the other hand, collecting low resolution labels (labels for a block of pixels) for these high resolution images is much more cost efficient. Conventional methods trained on these low-resolution labels are only capable of giving low-resolution predictions. The existing state-of-the-art label super resolution (LSR) method is capable of predicting high resolution labels, using only low-resolution supervision, given the joint distribution between low resolution and high resolution labels. Note that the set of low-resolution classes may differ from the set of high-resolution classes (\eg, 10 low-resolution classes reflecting the probability of presence of cancer, vs. 2 high-resolution classes of cancer or not). One major drawback of this existing method is that it does not consider the inter-instance variance which is crucial in the ideal mathematical formulation. In this work, we propose a novel loss function modeling the inter-instance variance. We test our method on a real world application: infiltrating breast cancer region segmentation in histopathology slides.  Experimental results show the effectiveness of our method.
\end{abstract}

\section{Introduction}

Given an input image $X=\{x_{i,j}\}$ with pixels $x_{i,j}$, a semantic segmentation model \cite{long2015fully,he2017mask,chen2018deeplab,papandreou2015weakly,badrinarayanan2017segnet,noh2015learning,havaei2017brain,hou2019sparse,xu2016deep} outputs a prediction image $Y=\{y_{i,j}\}$, where each $y_{i,j}$ is one of $L$ predefined classes: $y_{i,j}\in \left \{ 1, ...,L \right \}$.

\begin{figure}[ht]
\begin{center}
\includegraphics[width=0.85\linewidth]{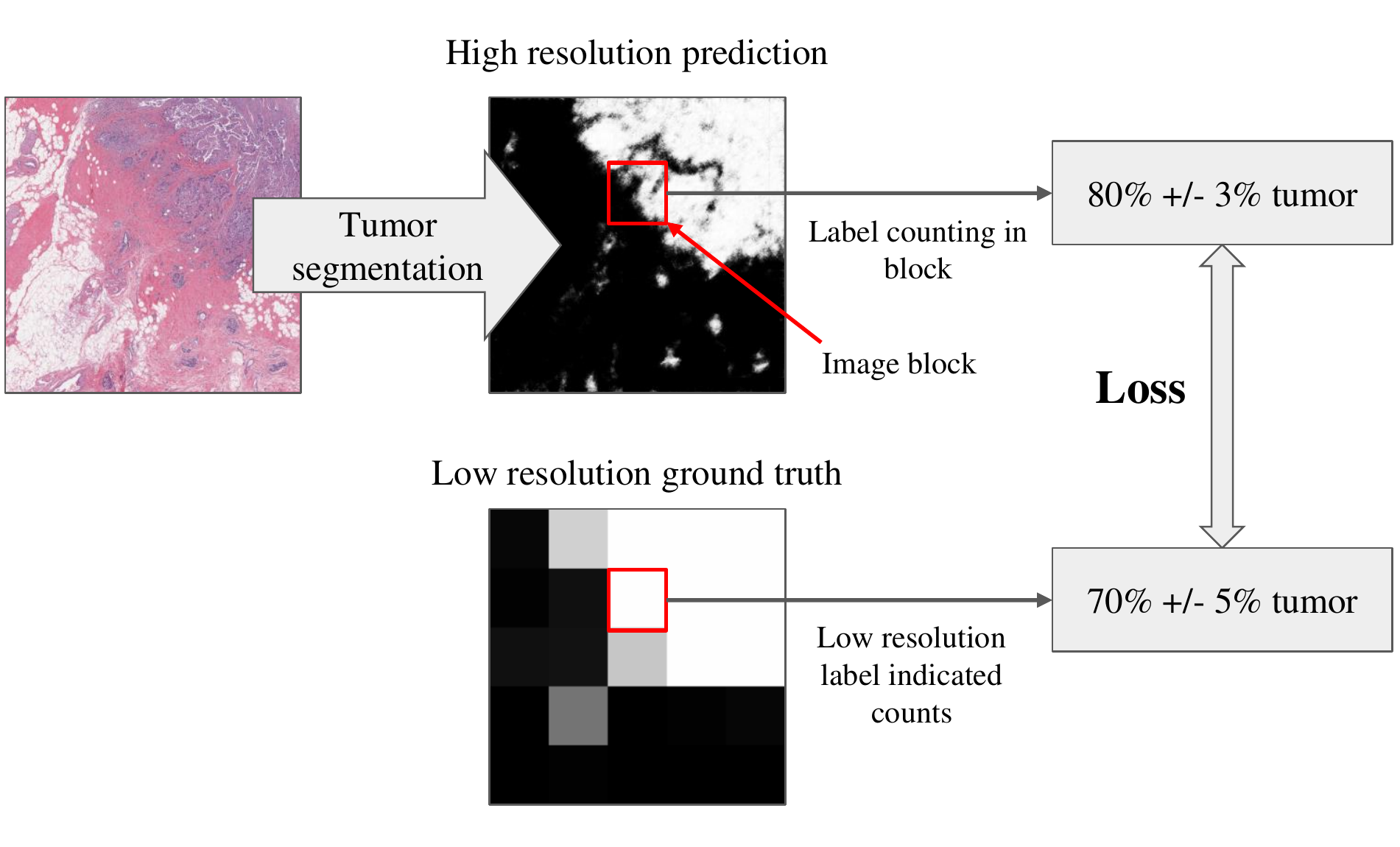}
\end{center}
   \caption{We focus on the problem of training a neural network for high-resolution semantic segmentation with low-resolution ground truth. The key component is to construct a loss between two distributions: predicted label count, and suggested label count from a low-resolution image block.}
\label{fig:concept}
\end{figure}

Conventional high-resolution semantic segmentation models require large amounts of high-resolution ground truth data (pixel-level labels)  \cite{badrinarayanan2017segnet,noh2015learning,havaei2017brain}. It is very labor intensive to collect these large scale datasets, especially for datasets of gigapixel images such as pathology images \cite{hou2016patch,liu2017detecting}. The weakly supervised semantic segmentation approaches \cite{chen2018deeplab,papandreou2015weakly,pathak2015constrained,wei2017stc,rakelly2018few} learn to produce pixel-level segmentation results given sparse, \eg, image-level labels. It requires that the set of image-level classes must be the same as the pixel-level classes. For example, given that the image contains a cat, the network learns to segment the cat \cite{papandreou2015weakly}. In many applications, 
however, low-resolution (\eg, block-level) information may correlate with pixel-level labels in a more complex way \cite{malkin19label}. For example, a patch in a tissue image may be 
assigned a \textit{probability} of containing cancer tissue and may contain  
\textit{high/low} amounts of different types of cells \cite{malkin19label,saltz2018spatial}.

The Label Super 
Resolution (LSR) method \cite{malkin19label} models this problem by utilizing the joint distribution 
between low-resolution and high-resolution labels, as shown in Fig. \ref{fig:concept}.
The LSR model is trained with each low resolution label $z$ assigned to each group of 
pixels (\ie, an image block) $X$. Let  $c_l$ be the number of pixels with high-resolution class label $l$ in an image block,  LSR tries to match the the actual count of $c_l$ in prediction with the count distribution $p(c_l \mid  z)$ indicated by $z$.

For 
\textit{each fixed image block}, the LSR loss matches the distribution of 
predicted $c_l$ given by the network, with the distribution of $c_l$ 
designated by the low resolution 
label $z$: $p(c_l \mid  z)$. Note that the ground truth $p(c_l \mid  z)$ is computed across multiple image blocks
with the same low resolution label $z$. On the other hand, the distribution of 
predicted $c_l$ is computed on each fixed image block. In other words, the existing LSR loss does not consider variance across image blocks with the same $z$. 


To address this problem, we propose a new loss function. The proposed loss functions 
match the distribution of $c_l$ across a set of image blocks with the same label $z$ to 
the distribution suggested by the low resolution label $p(c_l \mid z)$. Mathematically, 
this models the true variance of class/label counts across image blocks, not just within an image 
block.

 
We evaluate the proposed loss function on infiltrating breast cancer region
segmentation in Hematoxylin and Eosin stained pathology images.
The experiment results show that both of the loss functions outperform the LSR loss function significantly.
%
%
 To summarize, our contribution are as follows:
 \begin{enumerate}
     \item A novel loss functions for label super resolution, which takes into account variance across image blocks with the same low-resolution label.
     \item A breast cancer region segmentation model. The model can produce accurate high-resolution cancer segmentation boundary with only low resolution supervision in the training phase.
 \end{enumerate}  

The rest of the paper is organized as follows. Sec. \ref{sec:label-super-resolution} introduces the proposed loss functions; Sec. \ref{sec:experiments} describes the detailed implementation of our method, with experiments on the breast cancer region segmentation task; Finally, Sec. \ref{sec:conclusions} concludes this paper.

\section{Label Super Resolution}
\label{sec:label-super-resolution}
The existing Label Super Resolution (LSR) approach \cite{malkin19label} proposed an intra-instance loss function with which it learns to super-resolve low resolution labels. The key source of information it utilizes is the conditional distribution $p(c_l \mid  z)$: the probability distribution of $c_l$ within an image block with low resolution label $z$. As an example, Tab. \ref{tab:cancer_mu_sigma} shows $p(c_l \mid  z)$ 
for each high-resolution label $l$ and low-resolution label $z$ for the cancer segmentation task. In this example, $z$ is a binary label indicating if an image block is a cancer block or not; $l$ is a binary label indicating if a pixel 
is a cancer pixel or not (i.e., if the pixel is in a cancer cell or not). The cancer probability of an image block is provided by 
a patch-level cancer classifier \cite{liu2017detecting,hou2016patch}. The values in Tab. \ref{tab:cancer_mu_sigma} were computed through manual annotation. 
For each label $z$, a domain expert examined 10 to 12 $240\times240-pixel$ image blocks with label $z$ and estimated $c_l$ for each image block. In total, the domain expert examined 100 to 120 image blocks, instead of painstakingly delineating the precise boundaries of small and large cancer and non-cancer regions in whole slide tissue images. The cost of annotation in LSR is very low compared with conventional per-pixel labeling.

\begin{table}[ht]
\centering
\begin{tabular}{ c | c c }
Image block with & \multicolumn{2}{c}{Count\% of } \\
low resolution class $z$: & \multicolumn{2}{c}{high resolution class $l$:} \\ \cline{2-3}
probability\% as cancer block & cancer & Non-cancer \\ \hline\hline
0-20  & $0.0 \pm 0.1$ & $100.0 \pm  0.1$ \\ \hline
 20-30 & $1.0 \pm 0.4$ & $99.0 \pm 0.4$ \\ \hline
 30-40 & $2.0 \pm 0.4$ & $98.0 \pm 0.4$ \\ \hline
 40-50 & $5.0 \pm 0.8$ & $95.0 \pm 0.8$ \\ \hline
 50-60 & $6.0 \pm 1.0$ & $94.0 \pm 1.0$ \\ \hline
60-70   & $8.0 \pm 1.0$ & $92.0 \pm 1.0$ \\ \hline
70-80 & $10.0 \pm 1.0$ & $90.0 \pm 1.0$ \\ \hline
 80-90  & $10.0 \pm 1.0$ & $90.0 \pm 1.0$ \\ \hline
 90-95 & $20.0 \pm 2.0$ & $80.0 \pm 2.0$ \\ \hline
 95-100& $70.0 \pm 5.0$ & $30.0 \pm 5.0$ \\ \hline
\end{tabular}
\vspace{0.2cm}
\caption{\label{tab:cancer_mu_sigma}The distribution of the count (in percentage) of high resolution labels $l$ in image blocks with low resolution labels $z$. For example, in an image block with $90\%$ to $100\%$ probability of being cancer block, there are $70\%$ (expectation) $\pm 5\%$ (standard deviation) cancer pixels. In this case, the cancer block probability is given by a low resolution cancer classifier.}
\end{table}

\begin{figure*}[ht]
\begin{center}
\includegraphics[width=0.99\linewidth]{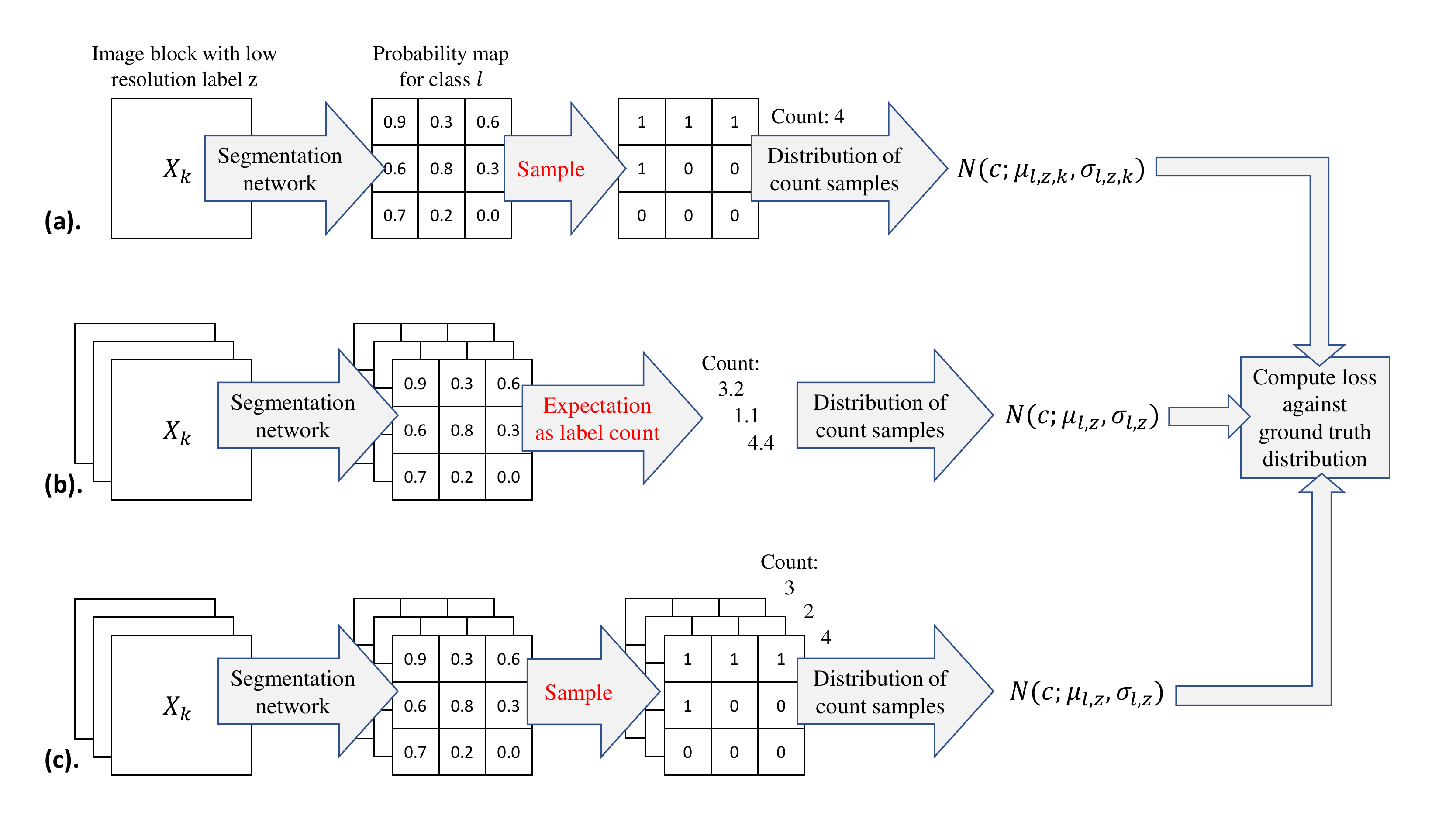}
\end{center}
   \caption{\textbf{(a).} The intra-instance loss baseline \cite{malkin19label} described in Sec. \ref{sec:intra-instance-loss}. This method models the label counts as a random variable, and derives the distribution of it given a \textit{fixed} input image block $X_k$. \textbf{(b).} Our proposed inter-instance loss described in Sec. \ref{sec:inter-instance-loss}. This method computes the distribution of label counts across input blocks, considering the label counts given by the network as a constant given a fixed input block. \textbf{(c).} Our proposed intra + inter-instance loss described in Sec. \ref{sec:intra-inter-instance-loss}. This method computes the distribution of label counts across input blocks, considering the label counts given by the network as a random variable given a fixed input image block.}
\label{fig:intra-inter-instance-loss}
\end{figure*}

All super resolution methods in this paper use the conditional distribution $p(c_l \mid  z)$. We first describe this baseline method \cite{malkin19label} as an intra-instance loss. We then formulate two new loss functions. An overview of these three loss functions is shown in Fig. \ref{fig:intra-inter-instance-loss}

\subsection{Baseline: intra-instance loss}
\label{sec:intra-instance-loss}
We introduce the intra-instance loss \cite{malkin19label} starting with \textit{label counting}. The classification/segmentation network produces, for each pixel in the image, 
a probability that a given pixel is in class $l$. This is expressed as $p_{\text{net}}(y_{i,j}=l \mid X_k, z)$, where $X_k$ is $k$-th input 
image block with low resolution label $z$; and $y_{i,j}$ is the class of 
a pixel with 
coordinates $i,j$. The LSR approach models the network's output on a pixel as a 
Bernoulli distribution. If we sampled the model's prediction at each pixel $i,j$, 
the value of $c_l$ would be 
\begin{equation}
    c_l = \frac{1}{| X_k |}\sum_{(i,j) \in X_k }\mathbbm{1} (y_{i,j}=l),
\end{equation}
where $\mathbbm{1}(\cdot)$ is the indicator function. Given a set of pixels in 
$X_k$, whose class label is $z$, the value of $c_l$ is approximated by a Gaussian distribution:
\begin{equation}
\label{eq:per-block-count}
    p_{\text{net}}(c_l = c \mid X_k, z) = N(c; \mu_{l,z,k}, \sigma_{l,z,k}^2) ,
\end{equation}
where
\begin{equation}
\label{eq:per-block-mu-sigma}
\begin{split}
\mu_{l,z,k}= & \frac{1}{| X_k |} \sum_{i,j}p(y_{i,j}=l\mid X_k, z), \\
\sigma_{l,z,k}^2= & \frac{1}{| X_k |} \sum_{i,j} \Big (p(y_{i,j}=l\mid X_k, z) \times \\ & \quad \big(1-p(y_{i,j}=l\mid X_k, z)\big)\Big ).
\end{split}
\end{equation}
As shown in Tab. \ref{tab:cancer_mu_sigma}, the ground truth is also modeled as a Gaussian distribution, only depending on the low resolution class $z$:
\begin{equation}
    p(c_l = c \mid z) = N(c; \eta_{l,z}, \rho_{l,z}^2).
\end{equation}

\paragraph{Statistics matching:} The LSR method minimizes the distance between $p_{\text{net}}(c_l = c \mid X_k, z)$ and $p(c_l = c \mid z)$ for each input 
$X_k$ with label $z$. The distance between two Gaussian distributions is formulated as follows:

\begin{equation}
\label{eq:match_loss}
D(p_{\text{net}},p)=-\textrm{log} \, p_{\text{net}}(C_l\mid I_z)
= \frac{1}{2} \frac{\sigma_{l,z,k}^2 (\eta_{l,z} - \mu_{l,z,k})^2 }{(\rho_{l,z} ^2+\sigma_{l,z,k}^2)^2} +\frac{1}{2}\textup{log}2\pi\sigma_{l,z,k}^2
\end{equation}

\paragraph{Drawback of Intra-instance Loss:} 
Given an instance (an image block) with a low-resolution label, the distribution 
of predicted class counts is computed by \textit{fixing} the input instance. 
In other words, $p_{\text{net}}(c_l = c \mid X_k, z)$ is computed instead of 
$p_{\text{net}}(c_l = c \mid z)$. By minimizing the dissimilarity between 
$p_{\text{net}}(c_l = c \mid X_k, z)$ and $p(c_l = c \mid z)$, a classification/segmentation network trained with the 
absolute optimal training error produces the same distribution of class counts 
$p_{\text{net}}(c_l = c \mid X_k, z)=p(c_l = c \mid z)$ \textit{regardless of 
the fact that $X_k$ is varied}.

\subsection{Inter-instance loss}
\label{sec:inter-instance-loss}


Because the distribution of real class counts is computed \textit{across} different instances (image blocks) with the same label $z$, we argue that one should also model the distribution of predicted class counts \textit{across} instances.

We formulate our proposed inter-instance loss as follows. First, we develop a new 
intra-instance loss. For each input instance $X_k$ with low resolution label $z$, 
the predicted value of $c_l$ is defined as the average predicted probability 
for high resolution class $l$. In this case, $p_{\text{net}}(c_l = c \mid X_k, z)$ 
is discrete:
\begin{equation}
p_{\text{net}}(c_l = c \mid X_k, z) =
\begin{cases}
      1 & \text{if }c=\mu_{l,z,k} ,\\
      0 & \text{otherwise.}
    \end{cases}
\end{equation}
In other words, we model the predicted count $c_l$ as a constant: $\mu_{l,z,k}$.

Using this simplified formulation, we model the predicted count $c_l$ 
across different instances $X_k$ as an approximate Gaussian distribution:
\begin{equation}
    p_{\text{net}}(c_l = c \mid z) = N(c;\mu_{l,z},\sigma_{l,z}^{2}),
\end{equation}
where $\mu_{l,z}$ and $\sigma_{l,z}$ are computed empirically:
\begin{equation}
\begin{split}
\mu_{l,z} = & \frac{1}{N}\sum_k \mu_{l,z,k}, \\
\sigma_{l,z}^{2} = & \frac{1}{N}\sum_k (\mu_{l,z,k}-\mu_{l,z})^2.
\end{split}
\end{equation}

In practice, it may not be possible to compute the exact $\mu_{l,z}$ and $\sigma_{l,z}^{2}$ when the number of image blocks $N$ is large and computational resources are limited. We address this problem by estimating $\mu_{l,z}$ and $\sigma_{l,z}^{2}$ on a batch of sample instances. This 
strategy is well in line with stochastic neural network training strategies.

The inter-instance loss is computed as follows:
\begin{equation}
\label{eq:inter-instance-statiscs-matching}
    D(p_{\text{net}},p)=-\textrm{log} \, p_{\text{net}}(C_l\mid I_z)
= \frac{1}{2} \frac{\sigma_{l,z}^2 (\eta_{l,z} - \mu_{l,z})^2 }{(\rho_{l,z} ^2+\sigma_{l,z}^2)^2} +\frac{1}{2}\textup{log}2\pi\sigma_{l,z}^2
\end{equation}

Our method matches $p_{\text{net}}(c_l = c \mid z)$ to $p(c_l = c \mid z)$ by assuming that the predicted value of $c_l$ is a constant given an input block $X_k$.

\paragraph{Drawback of Inter-instance Loss:} The inter-instance loss does not consider intra-image variation: the confidence of model prediction. Less confident predictions yield larger intra-image variations.



\subsection{Intra + inter-instance loss}
\label{sec:intra-inter-instance-loss}
Following the intra-instance loss formulation in Sec. \ref{sec:intra-instance-loss}, 
the predicted label counts vary when prediction for each pixel is viewed 
as a Bernoulli random variable.

Our intra+inter-instance loss is based on label count sampling. We have developed the following sampling strategy. Given low resolution label $z$, we first sample $X_k$. 
We then use the segmentation network to compute $p(y_{i,j}=l \mid X_k, z)$. 
Finally we sample a class count $c_l$ according to $p(y_{i,j}=l \mid X_k, z)$ for all $(i,j) \in X_k$. This across-block label count is approximated by the following Gaussian distribution:
%
%
\begin{equation}
p_{\text{net}}(c_l=c \mid z)=N(c;\mu_{l,z},\sigma_{l,z}^2).
\end{equation}
Here, $c_{l,z,k}$ is the label count, $c_l$, given $X_k$ with low resolution label $z$. We compute $\mu_{l,z}$ and $\sigma_{l,z}^{2}$ empirically:
\begin{equation}
\begin{split}
\mu_{l,z}= & E_k[\mu_{l,z,k}] \\
= & \frac{1}{N}\sum_{k=1}^{N}\left [ \frac{1}{| X_k |} \sum_{(i,j\in X_k)}p(y_{i,j}=l\mid X_k) \right ],\\
\sigma_{l,z}^{2}= & E_k\left [ \left (  c_{l,z,k}-\mu_{l,z}\right )^2 \right ]\\
= & E_k\left [   c_{l,z,k}^2\right ]+\mu_{l,z}^2-2\mu_{l,z}E_k[c_{l,z,k}] \\
= & \frac{1}{N}\sum_{k=1}^{N}\left (\sigma_{l,z,k}^2+\mu_{l,z,k}^2 \right )-\mu_{l,z}^2 \\
= & \frac{1}{N}\sum_{k=1}^{N}\sigma_{l,z,k}^2+ \frac{1}{N}\sum_{k=1}^{N}(\mu_{l,z,k}^2 -\mu_{l,z}^2) \text{.}
\end{split}
\end{equation}
We estimate $\mu_{l,z}$ and $\sigma_{l,z}^{2}$ using a batch of image blocks. We use Eq. \ref{eq:inter-instance-statiscs-matching} as the statistics matching loss.
\section{Experiments on breast cancer region segmentation}
\label{sec:experiments}

\begin{figure*}[t]
\begin{center}
\includegraphics[width=0.99\linewidth]{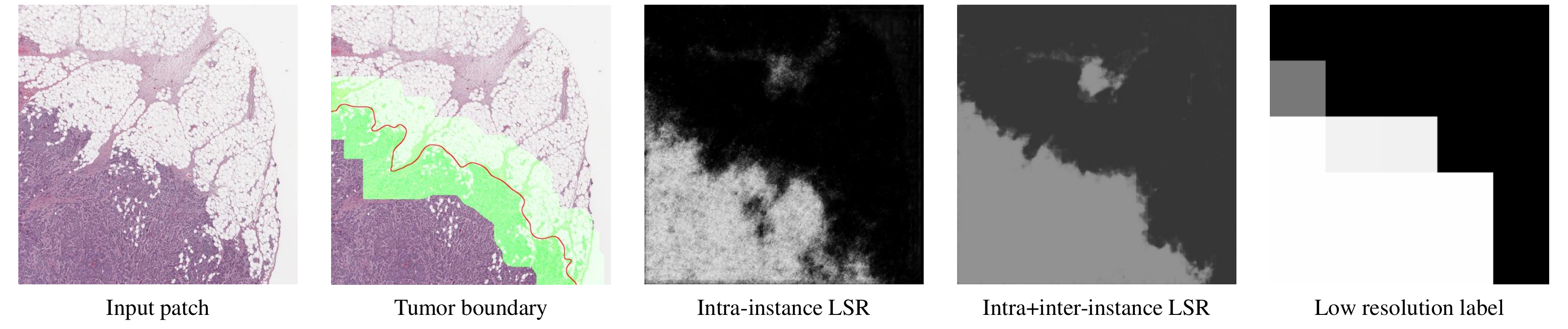}
\end{center}
   \caption{Examples of breast cancer segmentation results. The baseline intra-instance loss LSR \cite{malkin19label} generates prediction results with pepper noise, due to the lack of inter-instance variance modeling: it forces the network to predict a certain label count given a fixed low resolution label, regardless of its input image block (patch). On the other hand, our Intra+inter-instance yields smoother and more accurate segmentation results. The green area in the cancer boundary image indicates the mask in which we compute the masked IoU and DICE.}
\label{fig:cancer-visual-results}
\end{figure*}

Automatic cancer segmentation in pathology images has significant applications such as computer aided diagnosis and scientific studies \cite{gecer2018detection}. Manually annotating pixel-accurate cancer regions is time consuming, cost ineffective, and ambiguous. On the other hand, low resolution labels are relatively easy to collect and publicly available. Existing methods utilize low resolution labels to automatically produce low resolution segmentation results \cite{saltz2018spatial}. However, high resolution segmentation results have unique advantages such as showing accurate cancer boundaries which are important for the analysis of invasive carcinoma and infiltrating patterns of cancer \cite{jass1996assessment,weigelt2010molecular}. Our proposed method is able to produce high resolution segmentation results using the low-resolution annotations.

\subsection{Dataset}
We applied the proposed method to the task of cancer segmentation in breast carcinoma. Our low resolution labels are \textit{automatically generated} from a cancer/non-cancer region classifier. The cancer/non-cancer region classifier labels a patch of $4000\times4000$ pixels at a time, giving it a probability value of being cancer. The probability value is then quantified into 10 bins as 10 low resolution classes. Using this classifier, we labeled 1,092 breast carcinoma (BRCA) slides in The Cancer Genome Atlas (TCGA) repository \cite{TCGAdataset}, patch by patch. From 1,000 slides, we randomly extracted 26,767 patches with their low-resolution labels as training data. The patches from the rest 92 slides were for the validation and testing purposes. For training, the $4000\times4000$-pixel patches were downsampled to $240\times240$ pixels at 2.4X (4.2 microns per pixel). The classifier has a DICE score of 0.726 on the HASHI cancer segmentation dataset \cite{HASHI}, which has 196 TCGA slides. 

The joint distribution between the low resolution labels and the count of high resolution labels is in Tab. \ref{tab:cancer_mu_sigma}. 


\subsection{Evaluation method}
For evaluating our high resolution cancer segmentation results, we collected 
49 patches of $1200\times1200$ pixels at 2.5X magnification and 
carefully annotated cancer regions in detail. 42 of them are used as test set and 7 of them are used as validation set.
We use the Intersection over Union (IoU) and DICE coefficient scores as the evaluation metrics.

Since the only difference between low and high resolution cancer maps is reflected near cancer/non-cancer boundaries, we compute IoU and DICE scores only in areas within a distance of 240 pixels (1000 microns, width of an input patch) away from the ground truth cancer/non-cancer boundaries. We call those metrics as masked IoU and masked DICE. These two scores show performance difference only in regions that matter.

\subsection{Implementation details}
 We use a U-net-like architecture \cite{ronneberger2015u} with label super resolution losses. We do not use any high resolution data during training: only label super resolution methods are used. We use the RMSprop optimizer \cite{hinton2012neural} with $\beta=0.9$ to train all networks. In the intra-instance setting, we use a batch size of 30 and a learning rate of 0.00001. For the intra+inter-instance loss, the loss is computed using a group of 15 instances and each batch has 2 groups; and the learning rate is 0.001. 
\subsection{Experimental results}

We compare our methods to the original low resolution results given by the cancer/non-cancer region classification method. We call this the \textbf{low resolution model}. The quantitative results are shown in Tab. \ref{tab:cancer}. The proposed intra+inter-instance loss super resolves low resolution cancer region boundaries given by the low resolution model. This means that our method can generate finer cancer segmentation results with very limited amount of annotation labor overhead. More importantly, the network with the intra+inter-instance loss outperforms the network with the intra-instance loss.

\begin{figure}
\centering
\begin{tabular}{c c}
\begin{subfigure}[b]{0.38\textwidth}
\includegraphics[width=\textwidth]{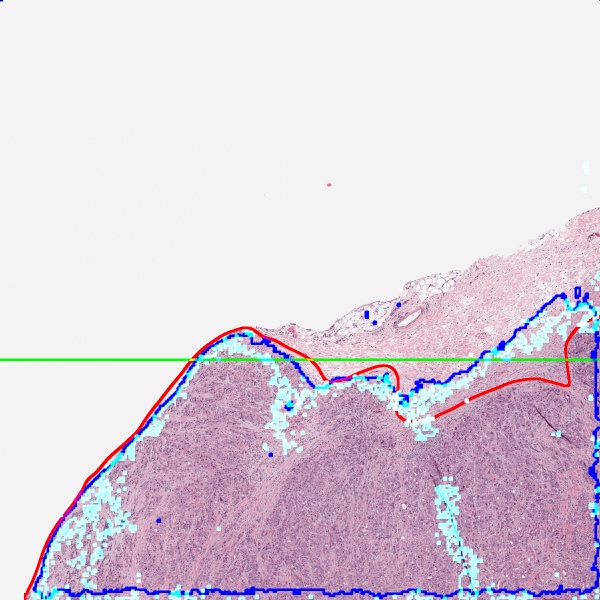}
\end{subfigure}&
\begin{subfigure}[b]{0.38\textwidth}
\includegraphics[width=\textwidth]{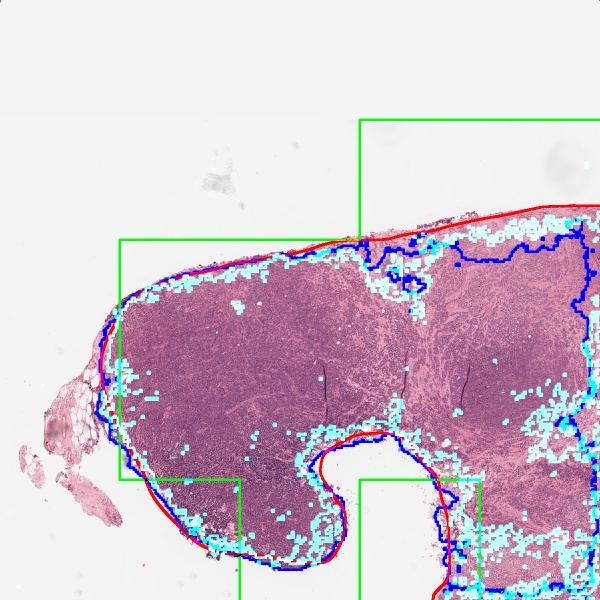}
\end{subfigure}\\
\begin{subfigure}[b]{0.38\textwidth}
\includegraphics[width=\textwidth]{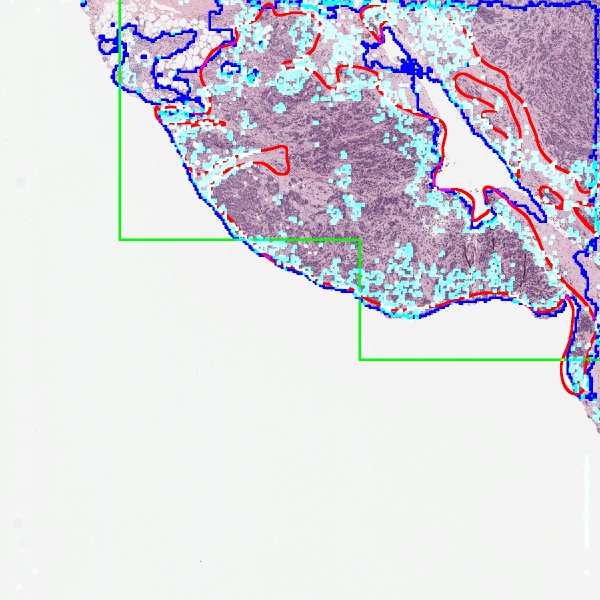}
\end{subfigure}&
\begin{subfigure}[b]{0.38\textwidth}
\includegraphics[width=\textwidth]{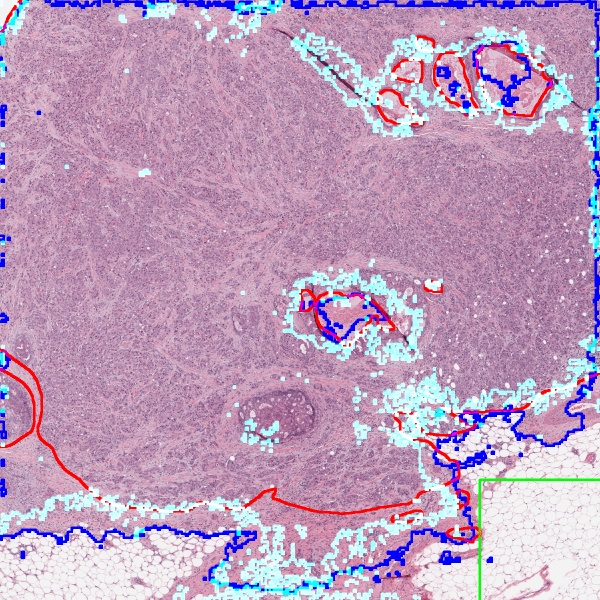}
\end{subfigure}\\
\begin{subfigure}[b]{0.38\textwidth}
\includegraphics[width=\textwidth]{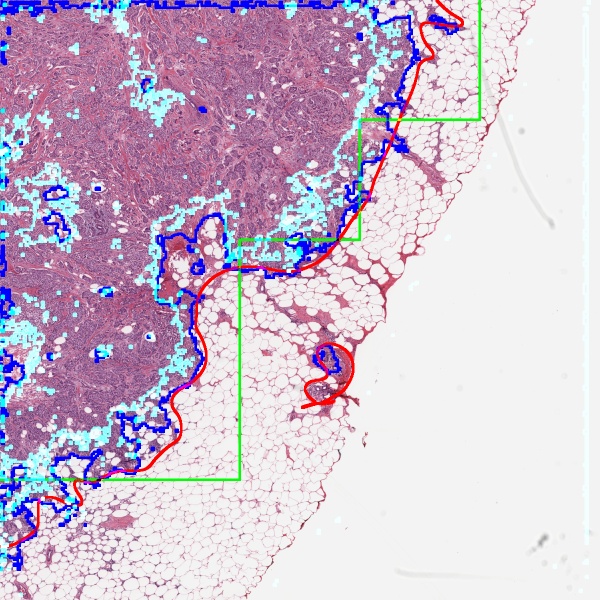}
\end{subfigure}&
\begin{subfigure}[b]{0.38\textwidth}
\includegraphics[width=\textwidth]{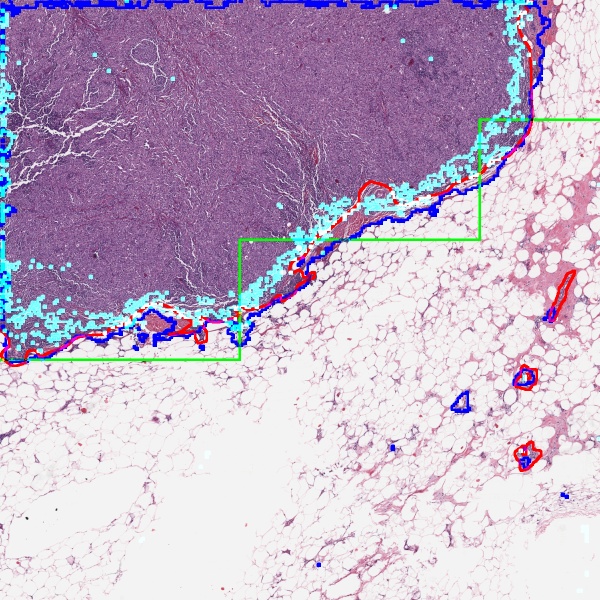}
\end{subfigure}\\
\end{tabular}
\caption{\label{fig:breast1}Visual examples of breast cancer segmentation. The green lines are the segmentation boundaries by thresholding the low resolution probability scores for patches . The red lines are ground truth cancer boundaries given by pathologists. The blue lines are the cancer segmentation boundaries predicted by the proposed Intra+inter-instance LSR. The cyan lines are the the cancer segmentation boundaries predicted by the Intra-instance LSR baseline.
   }
\end{figure}

Some qualitative results are in Fig. \ref{fig:cancer-visual-results} and Fig. \ref{fig:breast1}. From Fig. \ref{fig:cancer-visual-results} and Fig.  \ref{fig:breast1} we can see that, the baseline intra-instance loss LSR \cite{malkin19label} generates prediction results with pepper noise, due to the lack of inter-instance variance modeling: it forces the network to predict a certain label count given a fixed low resolution label, regardless of its input image block (patch). On the other hand, our Intra+inter-instance yields smoother and more accurate segmentation results. 

\begin{table}[]\centering 
\begin{tabular}{l c c}
            &  Masked  IoU & Masked DICE\\ \hline \hline
Low resolution model  &  0.5722 & 0.7278\\ \hline
Intra-instance    & 0.5810 & 0.7350\\ \hline
Intra+inter-instance & \textbf{0.5953} & \textbf{0.7463} \\ \hline
\end{tabular}
\vspace{0.2cm}
\caption{\label{tab:cancer}Quantitative results for cancer segmentation in pathology slides. The masked IoU/DICE is computed only in areas around cancer/non-cancer boundaries. It evaluates label super resolution methods in areas that matter, since prediction results totally inside/outside cancer regions do not need to be super resolved. In this sense, the proposed intra+inter-instance loss yields better results compared to the original low resolution cancer results. The network with intra+inter-instance loss outperforms the network with intra-instance loss consistently.
}
\end{table}

\section{Conclusions}
\label{sec:conclusions}
The high cost of high resolution annotations to train pixel-level 
classification and segmentation is a major roadblock to the effective 
application of deep learning in digital pathology and other domains 
that generate and analyze very high-resolution images. A label super 
resolution approach can address this problem by using low resolution 
annotations, but the current implementations do not take into account 
variations across image patches. The novel loss functions proposed in 
this work aim to alleviate this limitation. Our empirical results show 
that the across instance loss better captures and models the variance of high resolution labels within image blocks of the same low resolution label. As a result, they are capable of outperforming the existing baselines significantly. 
In the future, we plan to generalize this approach to detection networks, 
in addition to segmentation.

%
%
%

{\small
\bibliographystyle{ieee}
\bibliography{egbib}

\begin{thebibliography}{10}\itemsep=-1pt

\bibitem{badrinarayanan2017segnet}
V.~Badrinarayanan, A.~Kendall, and R.~Cipolla.
\newblock Segnet: A deep convolutional encoder-decoder architecture for image
  segmentation.
\newblock {\em IEEE transactions on pattern analysis and machine intelligence},
  39(12):2481--2495, 2017.

\bibitem{chen2018deeplab}
L.-C. Chen, G.~Papandreou, I.~Kokkinos, K.~Murphy, and A.~L. Yuille.
\newblock Deeplab: Semantic image segmentation with deep convolutional nets,
  atrous convolution, and fully connected crfs.
\newblock {\em IEEE transactions on pattern analysis and machine intelligence},
  40(4):834--848, 2018.

\bibitem{HASHI}
A.~Cruz-Roa, H.~Gilmore, A.~Basavanhally, M.~Feldman, S.~Ganesan, N.~Shih,
  J.~Tomaszewski, A.~Madabhushi, and F.~Gonz{\'a}lez.
\newblock High-throughput adaptive sampling for whole-slide histopathology
  image analysis (hashi) via convolutional neural networks: Application to
  invasive breast cancer detection.
\newblock {\em PloS one}, 13(5):e0196828, 2018.

\bibitem{gecer2018detection}
B.~Gecer, S.~Aksoy, E.~Mercan, L.~G. Shapiro, D.~L. Weaver, and J.~G. Elmore.
\newblock Detection and classification of cancer in whole slide breast
  histopathology images using deep convolutional networks.
\newblock {\em Pattern recognition}, 84:345--356, 2018.

\bibitem{havaei2017brain}
M.~Havaei, A.~Davy, D.~Warde-Farley, A.~Biard, A.~Courville, Y.~Bengio, C.~Pal,
  P.-M. Jodoin, and H.~Larochelle.
\newblock Brain tumor segmentation with deep neural networks.
\newblock {\em Medical image analysis}, 35:18--31, 2017.

\bibitem{he2017mask}
K.~He, G.~Gkioxari, P.~Doll{\'a}r, and R.~Girshick.
\newblock Mask r-cnn.
\newblock In {\em Proceedings of the IEEE international conference on computer
  vision}, pages 2961--2969, 2017.

\bibitem{hinton2012neural}
G.~Hinton, N.~Srivastava, and K.~Swersky.
\newblock Neural networks for machine learning lecture 6a overview of
  mini-batch gradient descent.
\newblock {\em Cited on}, 14, 2012.

\bibitem{hou2019sparse}
L.~Hou, V.~Nguyen, A.~B. Kanevsky, D.~Samaras, T.~M. Kurc, T.~Zhao, R.~R.
  Gupta, Y.~Gao, W.~Chen, D.~Foran, et~al.
\newblock Sparse autoencoder for unsupervised nucleus detection and
  representation in histopathology images.
\newblock {\em Pattern recognition}, 86:188--200, 2019.

\bibitem{hou2016patch}
L.~Hou, D.~Samaras, T.~M. Kurc, Y.~Gao, J.~E. Davis, and J.~H. Saltz.
\newblock Patch-based convolutional neural network for whole slide tissue image
  classification.
\newblock In {\em Proceedings of the IEEE Conference on Computer Vision and
  Pattern Recognition}, pages 2424--2433, 2016.

\bibitem{jass1996assessment}
J.~Jass, Y.~Ajioka, J.~Allen, Y.~Chan, R.~Cohen, J.~Nixon, M.~Radojkovic,
  A.~Restall, S.~Stables, and L.~Zwi.
\newblock Assessment of invasive growth pattern and lymphocytic infiltration in
  colorectal cancer.
\newblock {\em Histopathology}, 28(6):543--548, 1996.

\bibitem{malkin19label}
L.~H. R. S. J. C. D. S. J. S. L. J. N.~J. Kolya~Malkin, Caleb~Robinson.
\newblock Label super-resolution networks.
\newblock In {\em International Conference on Learning Representations (ICLR)},
  2019.

\bibitem{liu2017detecting}
Y.~Liu, K.~Gadepalli, M.~Norouzi, G.~E. Dahl, T.~Kohlberger, A.~Boyko,
  S.~Venugopalan, A.~Timofeev, P.~Q. Nelson, G.~S. Corrado, et~al.
\newblock Detecting cancer metastases on gigapixel pathology images.
\newblock {\em arXiv preprint arXiv:1703.02442}, 2017.

\bibitem{long2015fully}
J.~Long, E.~Shelhamer, and T.~Darrell.
\newblock Fully convolutional networks for semantic segmentation.
\newblock In {\em Proceedings of the IEEE conference on computer vision and
  pattern recognition}, pages 3431--3440, 2015.

\bibitem{noh2015learning}
H.~Noh, S.~Hong, and B.~Han.
\newblock Learning deconvolution network for semantic segmentation.
\newblock In {\em Proceedings of the IEEE international conference on computer
  vision}, pages 1520--1528, 2015.

\bibitem{papandreou2015weakly}
G.~Papandreou, L.-C. Chen, K.~P. Murphy, and A.~L. Yuille.
\newblock Weakly-and semi-supervised learning of a deep convolutional network
  for semantic image segmentation.
\newblock In {\em Proceedings of the IEEE international conference on computer
  vision}, pages 1742--1750, 2015.

\bibitem{pathak2015constrained}
D.~Pathak, P.~Krahenbuhl, and T.~Darrell.
\newblock Constrained convolutional neural networks for weakly supervised
  segmentation.
\newblock In {\em Proceedings of the IEEE international conference on computer
  vision}, pages 1796--1804, 2015.

\bibitem{rakelly2018few}
K.~Rakelly, E.~Shelhamer, T.~Darrell, A.~A. Efros, and S.~Levine.
\newblock Few-shot segmentation propagation with guided networks.
\newblock {\em arXiv preprint arXiv:1806.07373}, 2018.

\bibitem{ronneberger2015u}
O.~Ronneberger, P.~Fischer, and T.~Brox.
\newblock U-net: Convolutional networks for biomedical image segmentation.
\newblock In {\em International Conference on Medical image computing and
  computer-assisted intervention}, pages 234--241. Springer, 2015.

\bibitem{saltz2018spatial}
J.~Saltz, R.~Gupta, L.~Hou, T.~Kurc, P.~Singh, V.~Nguyen, D.~Samaras, K.~R.
  Shroyer, T.~Zhao, R.~Batiste, et~al.
\newblock Spatial organization and molecular correlation of tumor-infiltrating
  lymphocytes using deep learning on pathology images.
\newblock {\em Cell reports}, 23(1):181--193, 2018.

\bibitem{TCGAdataset}
{The TCGA team}.
\newblock {The Cancer Genome Atlas}.
\newblock \url{https://cancergenome.nih.gov/}.

\bibitem{wei2017stc}
Y.~Wei, X.~Liang, Y.~Chen, X.~Shen, M.-M. Cheng, J.~Feng, Y.~Zhao, and S.~Yan.
\newblock Stc: A simple to complex framework for weakly-supervised semantic
  segmentation.
\newblock {\em IEEE transactions on pattern analysis and machine intelligence},
  39(11):2314--2320, 2017.

\bibitem{weigelt2010molecular}
B.~Weigelt, F.~C. Geyer, R.~Natrajan, M.~A. Lopez-Garcia, A.~S. Ahmad,
  K.~Savage, B.~Kreike, and J.~S. Reis-Filho.
\newblock The molecular underpinning of lobular histological growth pattern: a
  genome-wide transcriptomic analysis of invasive lobular carcinomas and
  grade-and molecular subtype-matched invasive ductal carcinomas of no special
  type.
\newblock {\em The Journal of Pathology: A Journal of the Pathological Society
  of Great Britain and Ireland}, 220(1):45--57, 2010.

\bibitem{xu2016deep}
J.~Xu, X.~Luo, G.~Wang, H.~Gilmore, and A.~Madabhushi.
\newblock A deep convolutional neural network for segmenting and classifying
  epithelial and stromal regions in histopathological images.
\newblock {\em Neurocomputing}, 191:214--223, 2016.

\end{thebibliography}
}
\pagebreak

\end{document}